\pdfoutput=1

\documentclass[11pt]{article}

\usepackage{acl} 

\usepackage{times}
\usepackage{latexsym}

\usepackage[T1]{fontenc}

\usepackage[utf8]{inputenc}

\usepackage{microtype}

\usepackage{inconsolata}

\usepackage{fancyhdr}

\title{OTTAWA: Optimal TransporT Adaptive Word Aligner for Hallucination and Omission Translation Errors Detection}

\author{Chenyang Huang$^{1\dag}$\thanks{Equal contribution}\hspace{3mm}Abbas Ghaddar$^2$\footnotemark[1]\hspace{3mm}Ivan Kobyzev$^2$ \\
{\bf Mehdi Rezagholizadeh$^2$\hspace{3mm}Osmar R. Zaïane$^1$\hspace{3mm}Boxing Chen$^2$} \\
$^1$\normalsize{Dept. of Computing Science, Alberta Machine Intelligence Institute (Amii), University of Alberta}\\
$^2$Huawei Noah’s Ark Lab, Montreal Research Center, Canada\\
\texttt{\{chenyangh,zaiane\}@ualberta.ca}\\  
\texttt{\{abbas.ghaddar,ivan.kobyzev\}@huawei.com} \\
\texttt{\{mehdi.rezagholizadeh,boxing.chen\}@huawei.com}\\
}

\usepackage{booktabs}
\usepackage{eqparbox}
\usepackage{arydshln}
\usepackage{subcaption}
\usepackage{rotating,csquotes}
\usepackage{xcolor}
\usepackage{multirow}
\usepackage{framed}
\usepackage{booktabs}
\usepackage{amsmath}
\usepackage{amssymb}
\usepackage{mathtools}
\usepackage{upgreek}
\usepackage{soul}
\usepackage{amsfonts}
\usepackage{todonotes}
\usepackage{placeins}
\usepackage[most]{tcolorbox}
\usepackage{tabularx}
\usepackage{spverbatim}

\usepackage{amssymb,bm}
\usepackage{pifont}

\newtcbox{\mybox}[1][]{enhanced, colframe=blue, colback=blue!15, 
  frame style={opacity=0.25}, interior style={opacity=0.25}, 
  nobeforeafter, tcbox raise base, shrink tight, extrude by=1mm, #1}

\newcommand{\bert}{\textsc{Bert}}

\newcommand{\method}{\textsc{OTTAWA}}

\newcommand{\mightmention}[1]{}
\newcommand{\problem}[1]{\textcolor{red}{$\star$}}
\newcommand{\answer}[1]{\textcolor{blue}{$\#$}}
\newcommand{\todoreview}[1]{\textcolor{green}{$@$}}

\usepackage{mathtools}
\usepackage[thmmarks,amsmath]{ntheorem}

\usepackage{lastpage} 

\theoremnumbering{arabic}
\theoremstyle{break}
\theoremsymbol{}
\theoremheaderfont{\normalfont\bfseries}
\theorembodyfont{\upshape}

\theoremstyle{nonumberbreak}

\usepackage[most]{tcolorbox}
\newcommand{\specialcell}[2][c]{%
  \begin{tabular}[#1]{@{}c@{}}#2\end{tabular}}

\newcommand\blfootnote[1]{%
  \begingroup
  \renewcommand\thefootnote{}\footnote{#1}%
  \addtocounter{footnote}{-1}%
  \endgroup
}

\begin{document}
\maketitle
\begin{abstract}
\blfootnote{\!\!$^{\dag}$Work is done during an internship at Noah's Ark Lab.}
Recently, there has been considerable attention on detecting hallucinations and omissions in Machine Translation (MT) systems. The two dominant approaches to tackle this task involve analyzing the MT system's \textit{internal} states or relying on the output of \textit{external} tools, such as sentence similarity or MT quality estimators. In this work, we introduce \method{}, a novel Optimal Transport (OT)-based word aligner specifically designed to enhance the detection of hallucinations and omissions in MT systems. Our approach explicitly models the missing alignments by introducing a ``null'' vector, for which we propose a novel one-side constrained OT setting to allow an adaptive null alignment. Our approach yields competitive results compared to state-of-the-art methods across 18 language pairs on the HalOmi benchmark. In addition, it shows promising features, such as the ability to distinguish between both error types and perform word-level detection without accessing the MT system's internal states.\footnote{Our code is publicly available at \url{https://github.com/chenyangh/OTTAWA}} 
\end{abstract}

\section{Introduction}
Concerns regarding hallucination (as known as fabrication) of Machine Translation (MT) systems~\cite{raunak2021curious,muller2021understanding,guerreiro2023hallucinations} have led to considerable efforts towards the creation of diagnostic datasets~\cite{zhou2021detecting,guerreiro2023looking,dale2023halomi} and developing detection methods~\cite{guerreiro2023optimal,dale2023detecting}. In addition to hallucination errors, which occur when words in the translation are detached from the source sequence, addressing omission errors is also crucial—these are cases where words from the source sequence are absent in the translation.

\begin{figure}[t]
    \centering
    \includegraphics[width=0.95\linewidth]{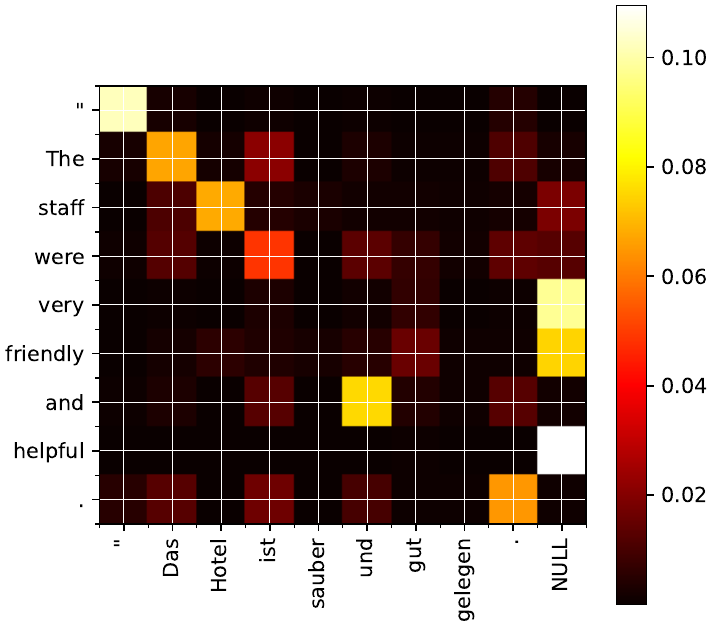} 
    \caption{A hallucinatory German-English translation was detected by correctly identifying the null alignments with our Optimal Transport-based method. Here, if a target word is mapping to ``NULL'', it is likely to be hallucinated, as no source word supports its translation.}
    \label{fig:hall_exp}
\end{figure}

Existing methods for detecting hallucinations and omissions in MT either focus on analyzing the model's \textit{internal} states (e.g. cross-attention) or rely on the output of \textit{external} tools, such as cross-lingual similarity estimators~\cite{feng2022language,heffernan2022bitext} or MT quality estimators~\cite{rei2020comet,duquenne2308sonar}. On one hand, \textit{internal} methods are limited to scenarios where white-box access to the MT system is assumed, and are further limited by dependencies on specific model architectures (e.g. attention-based). On the other hand, \textit{external} methods heavily rely on scalar outputs from models optimized for related tasks, such as MT quality or sentence similarity, but not directly optimized for hallucination detection. While hallucinations and omissions constitute low-quality translations, the opposite does not always hold. Furthermore, despite their effectiveness in detecting hallucinations and omissions, existing \textit{internal} and \textit{external} methods cannot distinguish between these two error types. This is because they frame the task as anomaly detection, relying on a single \textit{outlier score} to detect both error types. 

Cross-lingual word alignment~\cite{brown19mathematics,och2003systematic} aligns source to target tokens in parallel sentence pairs, and it provides unique insight for translation errors like hallucination and omission. Following the definition by \newcite{guerreiro2023looking}, hallucinated translation happens when the generated content is largely detached from the source text; therefore, we can measure hallucination by the amount of ``misaligned'' words in the translated text. Similarly, an omission error is likely if a significant amount of source words are not found in the translation.

Despite the strong relevance between word alignments and translation errors, alignment tools have not yet been widely applied for hallucination and omission detection. In particular, we observe that existing alignment approaches~\cite{sabet2020simalign,azadi2023pmi,arase2023unbalanced} are often based on the assumption that the translation is correct, such that most of the source or target words can be aligned with each other. This assumption, however, does not hold for scenarios where translation errors have occurred; hence, the direct application of existing word alignment tools does not work well for detecting translation errors.

In this paper, we utilize Optimal Transport problem~\cite{villani2009optimal,cuturi2013sinkhorn} and propose \method{}, an \underline{O}ptimal \underline{T}ranspor\underline{T} \underline{A}daptive  \underline{W}ord \underline{A}ligner that is specifically tailored to aligning words for the pathological cases where significant hallucination and omission errors are present.
Our approach explicitly models the missing alignments by introducing a ``null'' vector.
Specifically, we propose a novel one-side constrained Optimal Transport (OT) problem, which allows the null vector to be adaptively aligned to any number of words. Moreover, we set the null vector as a centric point, which is not biased during the computation of the OT problem.  We show an example of our OT-based method in Figure~\ref{fig:hall_exp}.

We conduct extensive experiments using the newly proposed HalOmi benchmark~\cite{dale2023halomi}, designed for hallucination and omission detection, across a total of 18 language pairs. Results show that our word aligner-based approach, equipped with \method{}, performs on par with state-of-the-art \textit{internal} and \textit{external} detection methods~\cite{guerreiro2023optimal,dale2023detecting}. In contrast to previous approaches, we show that our method can distinguish between hallucination and omission errors, while \method{} being crucial for performances compared to using existing word aligners.

\section{Related Work}
Hallucination remains a persistent concern in machine translation systems, which have recently gained increased attention in the field. \citet{zhou2021detecting} studied token-level hallucinations in machine translation on a Chinese-English dataset, which they constructed by randomly corrupting some tokens in the translation sentences. \citet{guerreiro2023looking}  developed a dataset of 3.4k naturally occurring German-English translations, manually annotated at both the sentence and token levels for hallucinations and omissions detection. HalOmi~\cite{dale2023halomi} expanded the language coverage for this task by manually annotating hundreds of translation pairs across 18 language pairs.

Methods for detecting hallucinations and omissions in machine translation can be broadly categorized into two types: \textit{internal}, which analyzes the translation model's own outputs and states, and \textit{external}, which relies on additional tools or data outside the model itself~\cite{dale2023detecting}. 
\citet{guerreiro2023optimal} developed a method that employs various optimal transport-based distances~\cite{Kantorovich1942,villani2009optimal} to evaluate the abnormality in a machine translation (MT) model's internal cross-attention distribution of a given translation, by making comparisons with those from a collection of high-quality translations. This method not only requires access to the internal states of the MT model but also necessitates a large collection of cross-attention maps from ground truth parallel sequences for the targeted language pair.

\citet{dale2023detecting} investigated the use of other internal MT model components, such as length-normalized sequence log-probability (Seq-Logprob) and the token-token layer-wise interaction framework (ALTI)~\footnote{Originally proposed by~\cite{ferrando2022towards} for interpreting MT systems}, as alternative abnormality estimators that do not require external translation data. In addition, the authors suggested directly using scores from \textit{external} MT  quality estimator models, such as COMET-QE ~\cite{rei2020comet}, or employing the cosine similarity between the source and translation as generated by cross-lingual sentence similarity models, including LASER~\cite{heffernan2022bitext}, LaBSE~\cite{feng2022language}.

Word alignment~\cite{brown19mathematics, brown1993advances} has been extensively studied over the years in both cross-lingual~\cite{nagata2020supervised,dou2021word} and monolingual~\cite{nagata2020supervised,lan2021neural} settings. The task is approached with supervised learning methods~\cite{och2003systematic,ostling2016efficient}, which require sentence-level parallel data, as well as unsupervised methods that do not rely on such data. The latter approach involves leveraging token-token attention matrices~\cite{ghader2017does,zenkel2020end,zhang2021bidirectional} or pairwise similarity matrices between tokens' embedding vectors~\cite{sabet2020simalign,azadi2023pmi}, both extracted from a multilingual pre-trained language models~\cite{devlin2019bert,conneau2020unsupervised}. The application of Optimal Transport \cite[OT,][]{Monge1781,Kantorovich1942,villani2009optimal,cuturi2013sinkhorn} in natural language processing tasks was mainly focused on similarity estimation between textual segments~\cite{huang2016supervised,alqahtani2021using,lee2022toward,mysore2022multi}, and more recently for word alignment~\cite{azadi2023pmi,arase2023unbalanced}.

\section{Method}
We formulate the unsupervised word alignment task in \S~\ref{sec:Unsupervised Word Alignment}, and then provide an overview on how standard and partial optimal transport are employed to address the task in \S~\ref{sec:Standard Optimal Transport} and \S~\ref{sec:Partial Optimal Transport}, respectively. In \S~\ref{sec:method}, we describe our newly proposed optimal transport word aligner, \method{}, specifically adapted to enhance the detection of MT hallucinations and omissions.
 
\subsection{Unsupervised Word Alignment}
\label{sec:Unsupervised Word Alignment}
Given a source word sequence  $\bm s{=} [s_1, \dots, s_m]$ and a target word sequence $\bm t {=} [ t_1, \dots, t_n]$, let $E^{(s)} {=} [ \bm  e^{(s)}_1, \dots, \bm e^{(s)}_m ] \in \mathbb{R}^{m \times D}$ and $E^{(t)} {=} [\bm  e^{(t)}_1, \dots, \bm  e^{(t)}_n] \in \mathbb{R}^{n \times D}$ be the embedding matrices for the source and target sequences, respectively, where $D$ is the dimensionality of the embeddings. A cost matrix $C \in \mathbb{R}^{m \times n}$ is defined such that $C_{i,j}$ is the cosine distance between the $i$-th source vector and the $j$-th target vector. 

The goal is to compute the alignment matrix $\Gamma\in \{0,1 \}^{m\times n}$,  where $\Gamma_{i,j} = 1$ if the $i$-th source word is aligned to the $j$-th target word, such that the aligned words have the smallest distance in the cost matrix $C$. A straightforward approach to obtain $\Gamma$ involves making greedy decisions on each word alignment in both forward (source-to-target) and reverse (target-to-source) directions. Specifically, given the cost matrix $C$, the forward alignment $\Gamma_{i,j}^{(F)}$ can be obtained by:
\begin{equation}
    \Gamma_{i,j}^{(F)} = \begin{cases}
    1, & \text{if } j = \operatorname{argmin}_j C_{i,j}  \label{eq:fwd} \\
    0, & \text{otherwise}
    \end{cases}
\end{equation}
and similarly, the reverse alignment $\Gamma_{i,j}^{(R)}$ is computed by:
\begin{equation}
    \Gamma_{i,j}^{(R)} = \begin{cases}
    1, & \text{if } i = \operatorname{argmin}_i C_{i,j}  \label{eq:rev} \\
    0, & \text{otherwise}
    \end{cases}
\end{equation}

The final alignment matrix $\Gamma$ is obtained by taking the element-wise multiplication of $\Gamma^{(F)}$ and $\Gamma^{(R)}$. However, ~\citet{sabet2020simalign} found that greedy decisions tend to ignore the word pairs of relatively lower similarity. Therefore, the authors propose to formulate the word alignment as an assignment problem \cite{kuhn1955hungarian}, which consequently involves solving the following optimization problem:
\begin{align}
\begin{aligned}
  \Gamma^* = \underset{{\Gamma \in U^{(A)}}}{\arg\min} \sum_{i,j} C_{i,j} \Gamma_{i,j} \label{eq:binary_alignment}
\end{aligned}
\end{align}
where $U^{(A)} = \{ \Gamma \in \{0, 1\}^{m \times n}: \Gamma\mathbf{1}_n \le \mathbf 1_n, \Gamma^\top\mathbf{1}_m \le \mathbf 1_m,  \sum_{i,j} \Gamma_{i,j} = \min(m, n) \}$ is the set of all binary matrices with the sum of row and column equal to 1. The optimization in Eq.~\ref{eq:binary_alignment} can be solved by linear programming \cite{bourgeois1971extension}. 

\subsection{Standard Optimal Transport}
\label{sec:Standard Optimal Transport}
\newcite{chi2021improving} have noticed that binary assignments obtained by Eq.~\ref{eq:binary_alignment} pose challenges when aligning source and target texts of markedly different lengths. This is because the binary assignment assumes that each word in the shorter text corresponds to a word in the longer text, while the additional words in the longer text remain unaligned. To address this issue, \newcite{chi2021improving,dou2021word} leverage standard optimal transport as an alternative solution for word alignment. The goal is to compute a  matrix $P^* \in \mathbb{R}_+^{m\times n}$  as follows:
\begin{align}
\begin{aligned}
  P^* = \underset{{P \in U^{(O)}}}{\arg\min} \sum_{i,j} C_{i,j} P_{i,j} \label{eq:general_alignment}
\end{aligned}
\end{align}
where $U^{(O)} = \{ P \in \mathbb{R}_{+}^{m \times n} : P\mathbf{1}_n =  \bm \mu, P^\top \mathbf{1}_m = \bm \nu \}$, $\bm \mu = ( 1 / m, \dots, 1 / m)$, and $\bm \nu = (1 / n, \dots, 1 / n)$. The binary matrix $\Gamma$ can be obtained by replacing the cost matrix $C$ with the matrix $P^*$ in Eq.~\ref{eq:fwd} and Eq.~\ref{eq:rev}, followed by performing element-wise multiplication on the resultant matrices. It is important to note that standard OT strictly requires the use of marginal constraints on $\bm{\mu}$ and $\bm{\nu}$, which enforce the alignment of each source or target word. Therefore, computing forward and reverse alignments is crucial for performance, as these computations implicitly model null alignments. This necessity arises because standard OT-based word alignment assumes that there is a one-to-one mapping between each word in the source sequence and a corresponding word in the target sequence, and vice versa.

\subsection{Partial Optimal Transport}
\label{sec:Partial Optimal Transport}
The one-to-one mapping assumption of standard OT is not always applicable, particularly in tasks with unbalanced source-target word alignments, such as monolingual text summarization. In response, \newcite{arase2023unbalanced} introduce partial optimal transport~\cite{peyre2016gromov,chapel2020partial} word aligner that relaxes the one-to-one mapping assumption, explicitly allowing for source or target words to remain naturally unaligned. In their setting, the transport map $P$ can be found by solving the optimization problem as in Eq.~\ref{eq:general_alignment} but with different constraints:
\begin{align}
\begin{aligned}
  U^{(P)}  & = \{ P \in \mathbb{R}_{+}^{m \times n}  : P\mathbf{1}_n \le \bm \mu, \label{eq:partial_alignment} \\ 
  & P^\top \mathbf{1}_m \le \bm \nu,  \sum_{i,j} P_{i,j} = m \} 
\end{aligned}
\end{align}
where $\bm \mu $ and $\bm \nu$ are uniform distributions, $m \in (0, 1)$ is a hyperparameter controlling how many source and target words are aligned. Then, a threshold $\tau$ is used to obtain the binary alignment matrix $\Gamma$ from the matrix $P$. Although the partial OT-based drops the strong assumption of the standard OT, it requires extensive search over the 
hyperparameters $m$ and  $\tau$
to determine whether a source and target word are considered aligned or not. Furthermore, the conditions in Eq.~\ref{eq:partial_alignment} favor the alignment of closely similar words, because the marginal constraints for either side are not guaranteed. Consequently, this promotes a high recall for alignments at the expense of precision when source and target words are not highly similar.

\subsection{\method{}}
\label{sec:method}

Accurately identifying unaligned words in the target text is crucial for detecting MT hallucinations. Similarly, identifying unaligned words in the source text enhances the detection of omissions. We propose explicitly handle null alignments as a mapping to special \textit{null} words, denoted as $\mathrm{s}_{m+1}$ and $\mathrm{t}_{n+1}$, which are appended to the source $\mathbf{s}$ and target $\mathbf{t}$ sequences, respectively. We denote the embeddings corresponding to $\mathrm{s}_{m+1}$ and $\mathrm{t}_{n+1}$  as $\bm e_{{m+1}}^{(s)}$ and $\bm e_{{n+1}}^{(t)}$, respectively. Our approach addresses forward and reverse alignments separately, yet employs the same method for both. Therefore, we describe how $\Gamma^{(R)}$ is obtained, as $\Gamma^{(F)}$ can be derived analogously. It is important to note that in performing a reverse alignment, only $\bm{e}_{{m+1}}^{(s)}$ is appended to $E^{(s)}$, while $E^{(t)}$ remains unchanged, and vice versa for forward alignment.

Let $\bm e_{{m+1}}^{s}$ be denoted as $\bm e^{(\varnothing)}$ hereafter for simplicity. So far, $\bm e^{(\varnothing)}$ can be any vector but to make the OT problem meaningful, we need to 
impose constraints on the distance between $\bm e^{(\varnothing)}$ and the target word vectors $E^{(t)}$. Firstly, $\bm e^{(\varnothing)}$ should be equidistant to every target vector to avoid bias. In addition, these equal distances should be of the same scale
as the average pair-wise distances between the source and target vectors to avoid aggressive alignment to $e^{(\varnothing)}$.

To this end, we extend the cost matrix $C$ to $\bar C \in \mathbb{R}^{(m+1) \times n}$, where $\bar C_{i,j} = C_{i,j}$ for $i \in [1, m]$ and $\bar C_{m+1,j} = d$, where $d$ is the distance between $\bm e^{(\varnothing)}$ and the target word representations (by our assumption, it is the same for all target words). One can define $d$ in multiple ways. For example, a natural choice is to use the mean of pair-wise distances between the source and target vectors. However, we choose to use the median over the mean to avoid the influence of the outliers. It is also based on the intuition that $\bm e^{(\varnothing)}$ should serve as a \textit{centric point}, which is not biased during the computation of the optimal transport map.

For $\bm e^{(\varnothing)}$ to be realizable in the vector space $\mathbb{R}^D$, the equal distance $d$ has a lower bound $d_{\text{min}}$ (see Appendix~\ref{app:math}).
We finally define $d = \max(d_{\text{min}}, c)$, where $c$ is the median of all pair-wise distances between the source and target words. We reformulate  the optimization problem for the transport map in Eq.~\ref{eq:general_alignment} by replacing the cost matrix $C$ with $\bar{C}$ and relaxing the constraints as follows:
\begin{align}
\begin{aligned}
  U^{(\varnothing, R)}   = \{ & P \in \mathbb{R}_{+}^{(m+1) \times n} : P\mathbf{1}_n \le \bm \mu',  \\
  & P^\top \mathbf{1}_{m+1} = \bm \nu \} \label{eq:our_alignment}
\end{aligned}
\end{align}
where $\bm \mu' = (\bm \mu, 1)$ allows $\bm e^{(\varnothing)}$ to have a maximum marginal of 1, and to be aligned with any number of target vectors. 
We denote the 
solution of this optimization problem
as $P^{(\varnothing,R)}$. 
The reverse alignment matrix $\Gamma^{(R)}$ is computed as in Eq.~\ref{eq:rev} by replacing $C$ with $P^{(\varnothing,R)}$. The forward alignment matrix $\Gamma^{(F)}$ is computed analogously. 
The final alignment matrix $\Gamma$ is obtained by taking the element-wise product between $\Gamma^{(F)}$ and $\Gamma^{(R)}$. 

Our optimization problem in the set of Eq.~\ref{eq:our_alignment} is a special case of partial optimal transport, it combines the properties of Eq.~\ref{eq:general_alignment}  and Eq.~\ref{eq:partial_alignment}. For example, in the computation of $\Gamma^{(R)}$, we relax the marginal constraint on the source side, a necessity given that the marginal for $\bm e^{(\varnothing)}$ can range between 0 and 1. As a result, the model has the freedom to adaptively decide whether or not to make \textit{null alignment} based on our defined median distance $d$. Conversely, the marginal constraint on the target side remains unchanged, ensuring that each target vector is compared with the centric point $\bm e^{(\varnothing)}$. 

To apply word alignment for hallucination and omission detections, we propose to analyze the number of non-aligned words from the final alignment matrix $\Gamma$. For both source and target sentences, we obtain the ratio of the misaligned words, given by:
\vspace{-2mm}
\begin{align}
  r^{(R)} & = \frac{1}{m} \sum_{i=1}^{m} \mathbb{I}\left(\sum_{j=1}^{n} \Gamma_{i,j} > 0\right) \label{eq:hall_align_ratio}  \\  
  r^{(F)} & = \frac{1}{n} \sum_{j=1}^{n} \mathbb{I}\left(\sum_{i=1}^{m} \Gamma_{i,j} > 0\right) \label{eq:omi_align_ratio}
\end{align}
where $\mathbb{I}(\cdot)$ is the indicator function. Further, we accumulate the transported mass to $\bm e^{(\varnothing)}$ as the confidence of missing alignments, given by:
\[
c^{(R)} = \frac{1}{n} \sum_{j=1}^{n} P^{(\varnothing,R)}_{m+1,j}  \,\,; ~~ \, c^{(F)} = \frac{1}{m} \sum_{i=1}^{m} P^{(\varnothing,F)}_{i,n+1}
\]
The combined score $r^{(R)} + c^{(R)}$  is used to detect hallucinations, and $r^{(F)} + c^{(F)}$ is for omissions.

\section{Experimental Setup}
\subsection{Datasets and Evaluation Metrics}
We conducted experiments using the recently proposed HalOmi benchmark \cite{dale2023halomi}, which contains manually annotated data for MT hallucination and omission detection across 18 language pairs. It is constructed by pairing English sentences with translations in 5 high-resource languages (ar:Arabic, ru:Russian, es:Spanish, de:German, and zh:Mandarin) and 3 low-resource languages (ks:Kashmiri, mni:Manipuri, and yo:Yoruba), and a one zero-shot\footnote{The MT system utilized for generating the data was not trained on Spanish-Yoruba parallel corpora.} Spanish-Yoruba pair. Each language pair includes between 145 to 197 examples, totaling 2,865 overall. Each example is annotated with 4 labels: \textit{no}, \textit{small}, \textit{partial}, and \textit{full} hallucination, with a similar scheme applied for omissions. Therefore, we utilized the multi-class ROC AUC variant, as defined in HalOmi~\cite{dale2023halomi}.

\subsection{Baselines and Models}
We conduct experiments to compare \textbf{internal} and \textbf{external} approaches for detecting hallucinations and omissions in MT, alongside our newly proposed \textbf{word aligner}-based approach. More precisely, consider Seq-Logprob, ALTI, ALTI\textsuperscript{T}, and ATT-OT (originally proposed by ~\cite{guerreiro2023optimal}). Similarly, for external approaches, we report results from MT quality estimators LaBSE~\cite{feng2022language} and BLASER-QE~\cite{barrault2023seamlessm4t}, as they are the best performers in this category according to ~\cite{dale2023halomi}. We refer the readers to~\cite{dale2023halomi} for a detailed of these baselines. 

\begin{table*}[!ht] 

\centering
 \renewcommand{\arraystretch}{1.25}
\resizebox{1\textwidth}{!}{%
\begin{tabular}{l@{\hspace{0.75\tabcolsep}}l@{\hspace{0.75\tabcolsep}}|c@{\hspace{0.75\tabcolsep}}c@{\hspace{0.75\tabcolsep}}c@{\hspace{0.75\tabcolsep}}c@{\hspace{0.75\tabcolsep}}c@{\hspace{0.75\tabcolsep}}c@{\hspace{0.75\tabcolsep}}c@{\hspace{0.75\tabcolsep}}|c@{\hspace{0.75\tabcolsep}}c@{\hspace{0.75\tabcolsep}}c@{\hspace{0.75\tabcolsep}}c@{\hspace{0.75\tabcolsep}}c@{\hspace{0.75\tabcolsep}}c@{\hspace{0.75\tabcolsep}}c}  
\toprule  
\multirow{2}{*}{\textbf{\colorbox{blue!0}{Source}}} &
\multirow{2}{*}{\textbf{\colorbox{blue!0}{Lang}}} & 
\multicolumn{7}{c|}{\textit{Hallucination}} & \multicolumn{7}{c}{\textit{Omission}} \\ 
& & 
\textbf{\colorbox{purple!20}{{$\mathcal{I}1$}}}  &
\textbf{\colorbox{purple!20}{{$\mathcal{I}2$}}}&
\textbf{\colorbox{purple!20}{{$\mathcal{I}3$}}} &
\textbf{\colorbox{purple!20}{{$\mathcal{I}4$}}}&
\textbf{\colorbox{green!20}{{$\mathcal{E}1$}}} &
\textbf{\colorbox{green!30}{{$\mathcal{E}2$}}} &
\textbf{\colorbox{yellow!30}{{Our}}} &

\textbf{\colorbox{purple!20}{{$\mathcal{I}1$}}}  &
\textbf{\colorbox{purple!20}{{$\mathcal{I}2$}}}&
\textbf{\colorbox{purple!20}{{$\mathcal{I}3$}}} &
\textbf{\colorbox{purple!20}{{$\mathcal{I}4$}}}&
\textbf{\colorbox{green!20}{{$\mathcal{E}1$}}} &
\textbf{\colorbox{green!30}{{$\mathcal{E}2$}}} &
\textbf{\colorbox{yellow!30}{{Our}}} 
\\
\midrule 

\multicolumn{1}{l}{\multirow{10}{*}{\begin{sideways}\specialcell{High Resource}\end{sideways}}}      & en-ar &          89 &         78 &    54 &      36 &         84 &         90 & \textbf{93} &          63 &   44 &         76 &         64 &         79 & \textbf{85} &         79 \\
 & ar-en &          83 &         75 &    72 &      51 &         88 & \textbf{94} &         90 &          56 &   49 &         82 &         72 & \textbf{83} &         77 &         70 \\
 & en-ru &  \textbf{95} &         36 &    37 &      61 &         86 &         89 &         91 &          53 &   56 &         79 &         75 & \textbf{85} &         84 &         \textbf{85} \\
 & ru-en &          86 &         76 &    56 &      55 &         83 &         92 & \textbf{99} &          59 &   52 &         86 &         76 &         76 &         80 & \textbf{89} \\
 & en-es &          87 &         85 &    69 &      59 & \textbf{88} &         85 &         \textbf{88} &          39 &   31 & \textbf{89} &         89 &         86 &         80 &         87 \\
 & es-en &          92 &         89 &    67 &      37 & \textbf{94} &         87 &         87 &          61 &   50 &         59 &         66 & \textbf{73} &         71 &         \textbf{73} \\
 & en-de &          85 & \textbf{97} &    69 &      55 &         \textbf{97} &         87 &         88 &          50 &   46 &         77 &         74 &         66 & \textbf{85} &         81 \\
 & de-en &          90 &         80 &    59 &      65 &         95 & \textbf{97} &         96 &          48 &   38 &         82 & \textbf{83} &         70 &         70 &         74 \\
 & en-zh &  \textbf{88} &         82 &    47 &      60 &         86 &         78 &         \textbf{88} &          60 &   51 &         73 &         67 &         69 &         88 & \textbf{92} \\
 & zh-en &  \textbf{89} &         88 &    65 &      46 &         88 &         87 &         86 &          73 &   61 & \textbf{77} &         64 &         75 &         73 &         \textbf{77} \\

\midrule
\multicolumn{2}{l|}{\textbf{Avg. High Resource}} &          88 &         78 &    59 &      53 &         89 &         89 & \textbf{91} &          54 &   46 &         78 &         74 &         76 &         80 & \textbf{81} \\

\midrule

\multicolumn{1}{l}{\multirow{6}{*}{\begin{sideways}\specialcell{Low Resource}\end{sideways}}}  & en-ks &          68 &         71 &    74 &      54 &         76 & \textbf{81} &         78 &          50 &   52 & \textbf{90} &         81 &         76 &         77 &         84 \\
 & ks-en &          59 &         67 &    65 &      65 &         57 & \textbf{73} &         56 &          36 &   50 &         64 &         63 &         59 &         45 & \textbf{70} \\
& en-mni &          68 &         81 &    49 &      54 &         80 & \textbf{83} &         59 &          45 &   52 &         80 &         73 & \textbf{82} &         80 &         69 \\
& mni-en &          70 &         64 &    49 &      56 &         56 & \textbf{78} &         64 &          42 &   33 &         68 &         63 &         52 &         74 & \textbf{77} \\
 & en-yo &          77 &         74 &    59 &      65 &         56 &         79 & \textbf{80} &          77 &   50 &         70 &         72 &         65 &         67 & \textbf{86} \\
 & yo-en &          78 &         72 &    54 &      43 &         66 & \textbf{80} &         65 &  \textbf{68} &   60 &         65 &         55 &         66 &         56 &         \textbf{68} \\

\midrule
\multicolumn{2}{l|}{\textbf{Avg. Low Resouce}} &          70 &         71 &    57 &      56 &         66 & \textbf{79} &         67 &          53 &   49 &         73 &         68 &         67 &         67 & \textbf{76} \\

\midrule
\multirow{2}{*}{\parbox{1mm}{Zero-Shot}}  & yo-es &          60 & \textbf{65} &    47 &      44 &         56 &         57 &         54 &          62 &   47 & \textbf{85} &         69 &         62 &         66 &         76 \\
 & es-yo &          61 &         66 &    52 &      55 &         66 & \textbf{68} &         65 &          68 &   50 & \textbf{83} &         60 &         69 &         67 &         80 \\

\midrule
\multicolumn{2}{l|}{\textbf{Avg. Zero-Shot}} &          60 & \textbf{66} &    49 &      49 &         61 &         63 &         59 &          65 &   48 & \textbf{84} &         65 &         66 &         67 &         78 \\
\midrule
\multicolumn{2}{l|}{\textbf{Avg. HalOmi}} &          79 &         75 &    57 &      53 &         78 & \textbf{83} &         79 &          56 &   48 &         77 &         70 &         72 &         74 & \textbf{79} \\
\bottomrule  

\end{tabular} 
}

\caption{Methods performances (ROC AUC) on hallucination and omission detection across HalOmi's high-resource, low-resource, and zero-shot sets. Bold entries describe the best results among all methods, which we categorize under 3 approaches: \colorbox{purple!20}{{\underline{$\mathcal{I}$}nternal}}, \colorbox{green!20}{{\underline{$\mathcal{E}$}xternal}}, and \colorbox{yellow!20}{{\underline{Our} word alignment}}. \colorbox{purple!20}{{$\mathcal{I}1$, $\mathcal{I}2$, $\mathcal{I}3$, $\mathcal{I}4$}}: Seq-Logprob, ALTI, ALTI\textsuperscript{T}, and ATT-OT. \colorbox{green!20}{{$\mathcal{E}1$, $\mathcal{E}2$}}: LaBSE and BLASER-QE MT quality estimators. \colorbox{yellow!20}{{Our}}: \method{} using LaBSE token embeddings. All scores are scaled within the range of $[0, 100]$ following~\cite{dale2023halomi}. }   
\label{tab:hall_tab}
\end{table*} 

\subsection{Implementation Details}

Our implementation is based on PyTorch~\cite{paszke2019pytorch}, and we use the POT library\footnote{\url{https://pythonot.github.io/}} for all the Optimal Transport-based methods. For the entropy-regulized O (including the standard OT and partial OT), there is a regularization parameter $\epsilon$ that controls the sparsity of the OT solution, where 0.1 is the default value. In our experiments, we set $\epsilon$ to a low value of $0.05$ to encourage the confidence of the sovler. However, we observed that the performance is not sensitive, but an overly small $\epsilon$ may lead to numerical instability.   

In all experiments, we adopt a word-level representation approach which is consistent with standard practices in word alignment task~\cite{sabet2020simalign,azadi2023pmi}. Specifically, we first generate token-level representations for each token in the sequence and establish a token-to-word index mapping. Then, we apply mean pooling to average the tokens representations corresponding to the same word, resulting in word-level representations. In our main experiment uses the representations of the last layer of LaBSE~\cite{feng2022language}. When experimenting with other models, we use the hidden representation of the 8th layer of mBERT~\cite{devlin2019bert} and XLMR~\cite{conneau2020unsupervised}, respectively, following \cite{dou2021word,chi2021improving,azadi2023pmi}. 

Our \method{}, PMIAlign, and the standard OT do need extra hyper-parameters to determine word alignments. However, for partial OT, we set $m = 0.5$ and tested a few thresholds. Specifically, we set $\tau$ to 0.1, 0.05, and 0.025 of the maximal value in marginal $\mu$ and $\nu$, given by $\max(1/m, 1/n)$, and used the best threshold 0.05 to report the scores.  To obtain the hallucination and omission, all baseline aligners use the reverse missing alignment ratio Eq.~\ref{eq:hall_align_ratio} to estimate hallucination and use Eq.~\ref{eq:omi_align_ratio} to obtain omission scores.

\section{Results}
\subsection{Sentence-Level Detection}
Table~\ref{tab:hall_tab} shows the performances of methods across 3 approaches for hallucination and omission detection across HalOmi's 18 language pairs: 4 methods representing the \textit{internal} approach, 2 for the \textit{external} approach, and 1 showcasing our newly introduced \textit{word alignment} approach. The latter approach employs our \method{} word aligner, which leverages LaBSE to obtain token-level representations.

Although our method performs best on only 5 out of 18 language pairs for hallucination and 7 out of 18 for omission, it outperforms other methods in terms of overall average performance. More precisely, we outperform all baselines in high-resource languages for both hallucination and omission detection. However, we 
underperform in low-resource languages (except for omission detection) and zero-shot language pairs. Overall, although we underperform by 4\% compared to BLASER-QE ($\mathcal{E}2$) on average hallucination detection across the 18 language pairs, we surpass that model by 5\% on omission. It is worth mentioning that BLASER-QE is a strong baseline, it is an MT quality estimator that leverages the SONAR~\cite{duquenne2308sonar} sentence embeddings.
It was carefully tuned on 123.4k source-translation pairs, derived from outputs of 24 MT systems and covering 62 language pairs, all manually labeled by humans for MT quality estimation. 

However, our method significantly improves upon LaBSE ($\mathcal{E}2$), another MT quality estimator. While we report close tight (1\% gain on HalOmi average) on hallucination, we 
significantly outperform it by 6\% on omission. These results demonstrate that LaBSE’s token-level representations contain more expressive information for detecting omissions compared to the scalar values generated for MT quality estimation.

Finally, we observe that our method provides consistent and balanced performance in detecting both hallucinations and omissions compared to \textit{internal} methods. For instance, although Seq-Logprob ($\mathcal{I}1$) matches our method on hallucinations, it significantly underperforms compared to us by 23\% on omissions. Similarly, ALTI\textsuperscript{T} ($\mathcal{I}3$) has the smallest gap (2\%) with our method on omission but performs poorly, achieving only 57\% compared to our method's 79\%, on hallucination.

\subsection{Ablation Study}

We conduct ablation studies on our approach, testing variants that modify its two key components: the \textbf{Embedding} vector representation model and the word \textbf{Aligner}. Table~\ref{tab:ablation} presents the average scores for both HalOmi high-resource (HR) and low-resource (LR) language pair clusters, along with the overall average. The complete results across the 18 language pairs are presented in Table~\ref{tab:abl_table} in Appendix~\ref{app:Results}.

\begin{table}[!htp]
    \begin{center}
    
\resizebox{\columnwidth}{!}{
\begin{tabular}{l|ccc|ccc}
\toprule
& \multicolumn{3}{c|}{\textit{Hallucination}} & \multicolumn{3}{c}{\textit{Omission}} \\
& \bf HR & \bf LR & \bf Avg & \bf HR & \bf LR & \bf Avg \\
\midrule
\bf Embedding & & & & & & \\
\hspace{3mm} m\bert{} &         87 &         56 &         73 &         \textbf{81} &         69 &         76 \\
\hspace{3mm} XLMR &         88 &         53 &         71 &         78 &         64 &         72 \\
\hspace{3mm} LaBSE & \textbf{91} & \textbf{67} & \textbf{79} &         \textbf{81} & \textbf{76} & \textbf{79} \\
\midrule
\bf Aligner & & & & & & \\
\hspace{3mm} SimAlign  &         47 &         55 &         51 &         70 &         62 &         66 \\
\hspace{3mm} PMIAlign &         81 &         61 &         72 & \textbf{82} &         71 &         78 \\
\hspace{3mm} OT  &         81 &         60 &         72 &         \textbf{82} &         74 &         \textbf{79} \\
\hspace{3mm} POT  &         64 &         52 &         57 &         73 &         60 &         68 \\
\hspace{3mm} \method{}  & \textbf{91} & \textbf{67} & \textbf{79} & 81 & \textbf{76} & \textbf{79} \\

\bottomrule
\end{tabular}

    } 
 \end{center} 
  
\caption{Hallucation and omission detection average AUC scores on HalOmi's high-resource (HR) and low-resource (LR) languages. \textbf{Embeddings} refer to the results obtained by testing \method{} with token embeddings from various models. \textbf{Aligner} are the results when employing different word aligners, all utilizing LaBSE embeddings.}

\label{tab:ablation}
\end{table}
\paragraph{Embedding} We conduct experiments with \method{}, utilizing embeddings derived from unsupervised pre-trained models such as mBERT~\cite{devlin2019bert} and XLM-R~\cite{conneau2020unsupervised}. These models are commonly used in cross-lingual word alignment~\cite {sabet2020simalign,azadi2023pmi}. We notice that \method{} with LaBSE representations significantly outperforms the best-unsupervised representation baseline (m\bert{}) by 6\% and 3\% on the overall HalOmi average score on hallucination and omission, respectively. This is expected given that LaBSE representations are more task-specific than those from unsupervised models, having been extensively fine-tuned for translation pair quality estimation using carefully curated data for low-resource languages.

Consequently, the performance gap between LaBSE and models such as m\bert{} is larger in low-resource languages (11\% and \%3 on omissions) compared to high-resource languages (4\% and 0\% on hallucinations).  Overall, the results are promising, demonstrating that hallucinations and omissions can be detected even without access to the MT system or a robust MT quality estimator. This approach serves as an acceptable alternative, particularly in high-resource settings. However, its performance in low-resource languages necessitates further investigation.

\paragraph{Aligner} We compared \method{} with state-of-the-art word aligner methods, including SimAlign~\cite{sabet2020simalign} (Itermax), PMIAlign~\cite{azadi2023pmi}, standard OT~\cite{chi2021improving,dou2021word}, and POT~\cite{arase2023unbalanced}, all utilizing LaBSE representations. 

We first observe that the standard alignment methods work reasonably well on detecting omissions but not hallucinations. This is an understandable outcome, which can be explained as follows: 1) the hallucinated translations often contain a significant amount of "detached" words. For example, the word "cat" may be translated to "dog". This is particularly problematic for standard alignment methods as they assume the word "cat" should have a corresponding word in the translation, resulting in aligning "cat" to "dog" nevertheless. 2) the omission errors in the HalOmi dataset are often early terminated translations. In this case, the translated words are still mapped to the source words, and the standard methods are still able to align the existing words and detect the omission errors.

However, only our method works well in the detection of both hallucinations and omissions, where there is a 7\% overall average improvement over both PMIAlign and standard OT in hallucination errors.
This outcome is consistent with our motivation that the common assumption among standard word-alignment tools, which is that a mapping exists between all source and target words, prevents them from effectively identifying all errors. 

\subsection{Distinguish Hallucination and Omission}

\begin{figure}[ht!]
    \centering
     \includegraphics[scale=0.5]{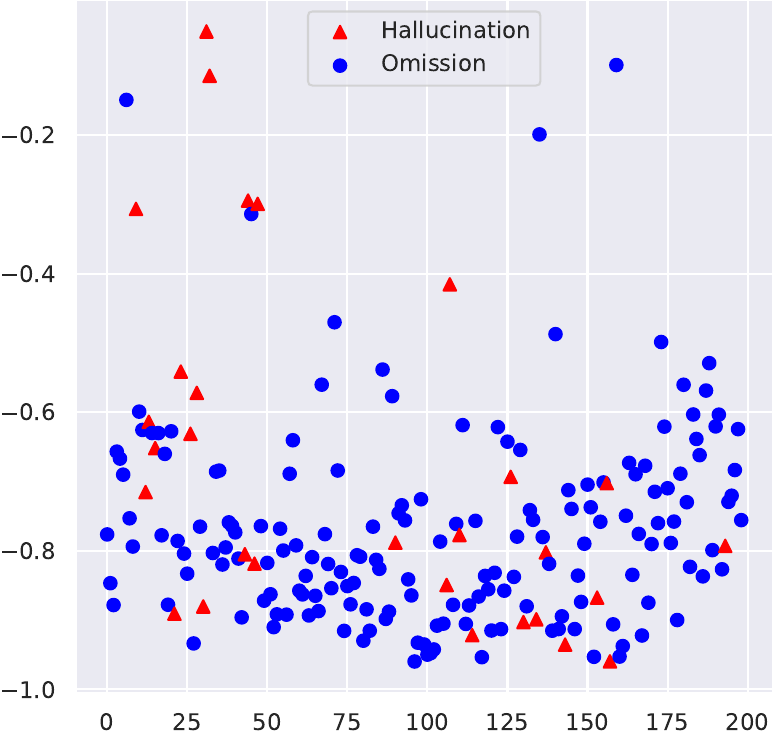} 
    
    \caption{LaBSE cosine similarity score (y-axis) of samples, aggregated across 10 high-resource datasets, with either hallucination (red triangles) or omission (blue circles) as gold labels. The x-axis shows the sample count index.}
    \label{fig:labase_plot}
\end{figure}

Figure~\ref{fig:labase_plot} shows LaBSE cosine similarity scores for samples with either hallucination or omission labels, aggregated from the 10 high-resource language pairs datasets. We only focus on high-resource languages, given that the performance of all methods significantly diminishes with low-resource ones. The primary goal of this experiment is to study the models' abilities to differentiate between types of errors. Although LaBSE quality estimator can effectively detect both hallucinations and omissions, the figure clearly shows that it fails to distinguish between them. The majority of samples for both hallucinations and omissions exhibit cosine similarity scores below $-0.8$.

\begin{figure}[ht!]
    \centering
     \includegraphics[scale=0.5]{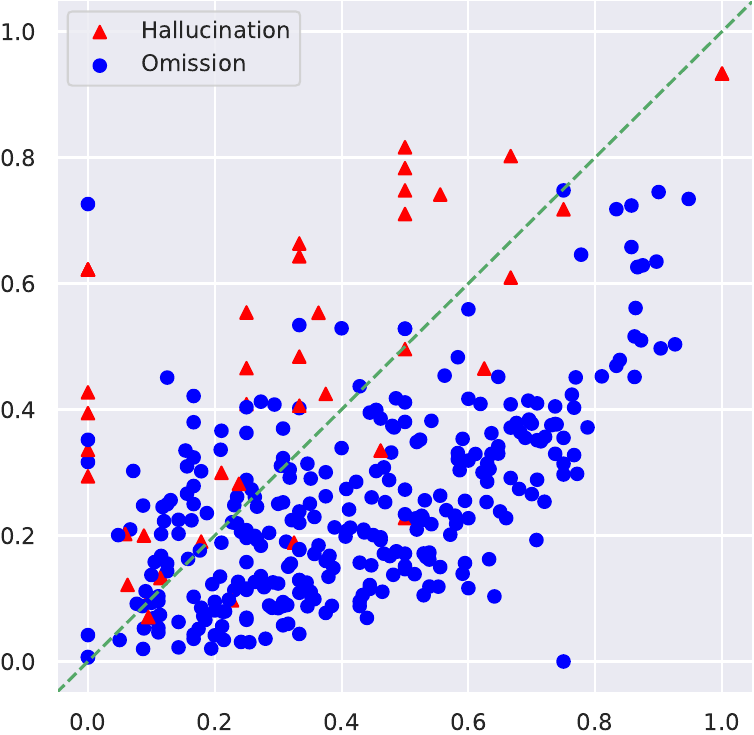} 
    
    \caption{Hallucination scores (x-axis) and Omission scores (y-axis) produced by our \method{} across the same samples used in Figure~\ref{fig:labase_plot}. The dotted green line simply illustrates the diagonal. }
    \label{fig:our_plot}
\end{figure}

Figure~\ref{fig:our_plot} shows the hallucination (x-axis)  and omission (y-axis) scores produced by \method{}(using LaBSE embeddings), on the same samples used in Figure~\ref{fig:labase_plot}.
The figure demonstrates that our approach effectively distinguishes between omission and hallucination errors, with omission errors concentrated below the diagonal, and hallucination errors concentrated above it. In most cases, our word alignment approach, which operates without any supervision signal, accurately distinguishes omissions from hallucinations, and vice versa. This unique feature offered by our approach can help researchers to precisely diagnose and comprehend error types in MT systems, thereby facilitating more efficient development and the implementation of targeted solutions.

\subsection{Word-Level Detection}

Following the experimental setup in \cite{dale2023halomi}, we evaluate our \method{} for word-level hallucination and omission detection. Table~\ref{tab:word_level} compares its performance with the best \textit{internal} approach baselines identified in their study: LogProb (the \texttt{Stand.} variant) and ALTI (the \texttt{sum} variant). It is worth noting that the LogProb and ALTI baselines are different from those in Table~\ref{tab:hall_tab} despite the name resemblance. Specifically, ALTI is a combination of two baselines from Table~\ref{tab:hall_tab}: ALTI (hallucination results), and ALTI\textsuperscript{T} (omission results). Conversely, the LogProb baseline is obtained by processing the sentence pairs through the MT model twice  (once for the source and once for the target), unlike SeqLogProb in Table~\ref{tab:hall_tab}, which involves only one pass. 

\begin{table}[!htp]
    \begin{center}
    
\resizebox{\columnwidth}{!}{
\begin{tabular}{l|ccc|ccc}
\toprule
& \multicolumn{3}{c|}{\textit{Hallucination}} & \multicolumn{3}{c}{\textit{Omission}} \\
& \bf HR & \bf LR & \bf Avg & \bf HR & \bf LR & \bf Avg \\
\midrule
\bf LogProb & 73 & 66 & 70 & 84 & \textbf{71} & 78  \\
\bf ALTI  & \textbf{87} & \textbf{69} & \textbf{78} & \textbf{86} & 69 & \textbf{78}  \\
\midrule
\bf \method{} &  78 & 59 & 70 & 79 & 60 & 71  \\
\bottomrule
\end{tabular}

    } 
 \end{center} 
  
\caption{Word-level hallucination and omission results, evaluated using the word-level ROC AUC score as defined by~\cite{dale2023halomi}.}

\label{tab:word_level}
\end{table}

Additionally, \cite{dale2023halomi} conducted ablations with these baselines, creating two variants for each: Standard and Contrastive for LogProb, and Mean and Max for ALTI. In Table~\ref{tab:word_level}, we report only the results for the best-performing variant of each baseline. Results indicate that word-level detection poses a greater challenge for our approach, which significantly under-performs compared to the best baseline (ALTI) by 8\% and 7\% in the overall average for hallucination and omission detection, respectively. However, we find the results encouraging when considering that unsupervised world-level detection with current state-of-the-art methods was not feasible without white-box access to the MT system, as \textit{external} approaches operate on sentence level only.

\subsection{Word Alignment}
\label{app: Word Alignment}

We run experiments on cross-lingual word alignment using 6 datasets, as in SimAlign~\cite{sabet2020simalign} and PMIAlign~\cite{azadi2023pmi}, with gold word alignment labels. We includes English-Czech (en-cs)~\cite{marevcek2011automatic}, German-English (de-en)~\cite{koehn2005europarl}, English-Persian (en-fa)~\cite{tavakoli2014phrase}, English-French (en-fr)~\cite{och2000improved}, English-Hindi (en-hi)~\cite{koehn2005proceedings} and Romanian-English (ro-en)~\cite{mihalcea2003evaluation} language pairs. For evaluation metrics, we utilize the standard Alignment Error Rate (AER)~\cite{och2003systematic}, and aligners use the m\bert{} representations. 

\begin{table}[!htp]
    \begin{center}
    
\resizebox{\columnwidth}{!}{
\begin{tabular}{lccccccc}
\toprule
& \bf en-cs & \bf de-en & \bf en-fa & \bf en-fr & \bf en-hi & \bf ro-en  \\ 

\midrule
\multicolumn{8}{c}{\textit{Alignment Error Rate (AER) {{ $\downarrow$}}}} \\
\midrule
SimAlign  & \bf 0.12 & 0.19 & 0.37 & 0.06 & 0.44 & 0.34  \\ 
PMIAlign  & \bf 0.12 & \bf 0.17 & \bf 0.32 & 0.06 & \bf 0.39 & 0.31  \\ 
OT  & \bf 0.12 & \bf 0.17 & \bf 0.32 & 0.06 & \bf 0.39 & 0.39  \\ 
POT  & 0.13 & 0.19 & 0.37 & 0.06 & 0.44 & 0.34  \\ 
\method{}  & \bf 0.12 & \bf 0.17 & 0.33 & \bf 0.05 & \bf 0.39 & \bf 0.30  \\ 
\bottomrule
\end{tabular}

    } 
 \end{center} 
  
\caption{Cross-lingual word alignment performances (AER) across 6 language pairs for 5-word aligners.}
\label{tab:word_align}
\end{table}

As shown in Table~\ref{tab:word_align}, \method{} performs on par with another state-of-the-art word aligner on the cross-lingual alignment task. More precisely, it matches the best-performing baselines in three languages (en-cs, de-en, and en-hi), while slightly outperforming them by 1\% in the en-fr and ro-en language pairs. We anticipated no significant gains in word alignment, as the datasets for cross-lingual word alignment are constructed from clean translations, making null alignments less concerning.

\section{Conclusion and Future Work}
In this work, we propose a new word alignment-based approach for detecting hallucinations and omissions in MT systems. We develop \method{}, an innovative word aligner designed specifically for this purpose. 
While experiments show promise in MT hallucination and omission detection, this area remains an intriguing direction for future research exploration. We plan to focus on one-to-many alignments for pathological translations.

\section{Limitations}
Our Optimal Transport (OT)-based word alignment method relies on pretrained word embeddings, which may not be available for low and scarce resource languages. Also, the performance of our word aligner on low resources heavily depends on representations obtained from a supervised MT quality estimator, which may not be accessible or does not support certain languages. Another limitation of our study is the exclusive focus on null alignments, which are central to our task of interest. However, we do not address other complex cases such as one-to-many and many-to-many alignments.   

\section*{Acknowledgements}
We thank the anonymous reviewers for their insightful comments. Osmar R. Za\"iane is supported by the Amii Fellow Program.

\bibliography{custom}

\begin{thebibliography}{50}
\expandafter\ifx\csname natexlab\endcsname\relax\def\natexlab#1{#1}\fi

\bibitem[{Alqahtani et~al.(2021)Alqahtani, Lalwani, Zhang, Romeo, and
  Mansour}]{alqahtani2021using}
Sawsan Alqahtani, Garima Lalwani, Yi~Zhang, Salvatore Romeo, and Saab Mansour.
  2021.
\newblock \href {https://aclanthology.org/2021.findings-emnlp.329} {Using
  optimal transport as alignment objective for fine-tuning multilingual
  contextualized embeddings}.
\newblock In \emph{Findings of the Association for Computational Linguistics:
  EMNLP 2021}, pages 3904--3919.

\bibitem[{Arase et~al.(2023)Arase, Bao, and Yokoi}]{arase2023unbalanced}
Yuki Arase, Han Bao, and Sho Yokoi. 2023.
\newblock \href {https://aclanthology.org/2023.acl-long.219} {Unbalanced
  optimal transport for unbalanced word alignment}.
\newblock In \emph{Proceedings of the 61st Annual Meeting of the Association
  for Computational Linguistics (Volume 1: Long Papers)}, pages 3966--3986.

\bibitem[{Azadi et~al.(2023)Azadi, Faili, and Dousti}]{azadi2023pmi}
Fatemeh Azadi, Heshaam Faili, and Mohammad~Javad Dousti. 2023.
\newblock \href {https://aclanthology.org/2023.findings-acl.782} {Pmi-align:
  Word alignment with point-wise mutual information without requiring parallel
  training data}.
\newblock In \emph{Findings of the Association for Computational Linguistics:
  ACL 2023}, pages 12366--12377.

\bibitem[{Barrault et~al.(2023)Barrault, Chung, Meglioli, Dale, Dong, Duquenne,
  Elsahar, Gong, Heffernan, Hoffman et~al.}]{barrault2023seamlessm4t}
Lo{\"\i}c Barrault, Yu-An Chung, Mariano~Cora Meglioli, David Dale, Ning Dong,
  Paul-Ambroise Duquenne, Hady Elsahar, Hongyu Gong, Kevin Heffernan, John
  Hoffman, et~al. 2023.
\newblock \href {https://arxiv.org/abs/2308.11596} {Seamlessm4t-massively
  multilingual \& multimodal machine translation}.
\newblock \emph{ArXiv Preprint}.

\bibitem[{Bourgeois and Lassalle(1971)}]{bourgeois1971extension}
F.~Bourgeois and J.~Lassalle. 1971.
\newblock \href {https://doi.org/10.1145/362919.362945} {An extension of the
  munkres algorithm for the assignment problem to rectangular matrices}.
\newblock \emph{Communications of the ACM}, 14(12):802--804.

\bibitem[{Brown et~al.(1993{\natexlab{a}})Brown, Della~Pietra, Della~Pietra,
  and Mercer}]{brown19mathematics}
Peter~E Brown, Vincent~J Della~Pietra, Stephen~A Della~Pietra, and Robert~L
  Mercer. 1993{\natexlab{a}}.
\newblock \href {https://www.aclweb.org/anthology/J93-2003} {The mathematics of
  statistical machine translation: Parameter estimation}.
\newblock \emph{Computational Linguistics}, 19(2).

\bibitem[{Brown et~al.(1993{\natexlab{b}})Brown, Della~Pietra, Della~Pietra,
  and Mercer}]{brown1993advances}
Peter~F Brown, Vincent~J Della~Pietra, Stephen~A Della~Pietra, and Robert~L
  Mercer. 1993{\natexlab{b}}.
\newblock \href {https://aclanthology.org/J01-3007.pdf} {Advances in
  probabilistic and other parsing technologies}.
\newblock \emph{IBM Journal of Research and Development}, 37(3):288--307.

\bibitem[{Campbell and Meyer(2009)}]{Campbell1979GeneralizedIO}
Stephen~L. Campbell and Carl~Dean Meyer. 2009.
\newblock \href {https://core.ac.uk/download/pdf/82174349.pdf}
  {\emph{Generalized inverses of linear transformations}}.
\newblock Society for Industrial and Applied Mathematics.

\bibitem[{Chapel et~al.(2020)Chapel, Alaya, and Gasso}]{chapel2020partial}
Laetitia Chapel, Mokhtar~Z. Alaya, and Gilles Gasso. 2020.
\newblock \href
  {https://proceedings.neurips.cc/paper/2020/hash/1e6e25d952a0d639b676ee20d0519ee2-Abstract.html}
  {Partial optimal transport with applications on positive-unlabeled learning}.
\newblock In \emph{Advances in Neural Information Processing Systems},
  volume~33, pages 9778--9789.

\bibitem[{Chi et~al.(2021)Chi, Dong, Zheng, Huang, Mao, Huang, and
  Wei}]{chi2021improving}
Zewen Chi, Li~Dong, Bo~Zheng, Shaohan Huang, Xian-Ling Mao, He-Yan Huang, and
  Furu Wei. 2021.
\newblock \href {https://aclanthology.org/2021.acl-long.265} {Improving
  pretrained cross-lingual language models via self-labeled word alignment}.
\newblock In \emph{Proceedings of the 59th Annual Meeting of the Association
  for Computational Linguistics and the 11th International Joint Conference on
  Natural Language Processing (Volume 1: Long Papers)}, pages 3418--3430.

\bibitem[{Conneau et~al.(2020)Conneau, Khandelwal, Goyal, Chaudhary, Wenzek,
  Guzm{\'a}n, Grave, Ott, Zettlemoyer, and Stoyanov}]{conneau2020unsupervised}
Alexis Conneau, Kartikay Khandelwal, Naman Goyal, Vishrav Chaudhary, Guillaume
  Wenzek, Francisco Guzm{\'a}n, {\'E}douard Grave, Myle Ott, Luke Zettlemoyer,
  and Veselin Stoyanov. 2020.
\newblock \href {https://aclanthology.org/2020.acl-main.747} {Unsupervised
  cross-lingual representation learning at scale}.
\newblock In \emph{Proceedings of the 58th Annual Meeting of the Association
  for Computational Linguistics}, pages 8440--8451.

\bibitem[{Cuturi(2013)}]{cuturi2013sinkhorn}
Marco Cuturi. 2013.
\newblock \href
  {https://proceedings.neurips.cc/paper_files/paper/2013/file/af21d0c97db2e27e13572cbf59eb343d-Paper.pdf}
  {Sinkhorn distances: Lightspeed computation of optimal transport}.
\newblock In \emph{Advances in Neural Information Processing Systems},
  volume~26.

\bibitem[{Dale et~al.(2023{\natexlab{a}})Dale, Voita, Barrault, and
  Costa-juss{\`a}}]{dale2023detecting}
David Dale, Elena Voita, Loic Barrault, and Marta~R. Costa-juss{\`a}.
  2023{\natexlab{a}}.
\newblock \href {https://aclanthology.org/2023.acl-long.3} {Detecting and
  mitigating hallucinations in machine translation: Model internal workings
  alone do well, sentence similarity {E}ven better}.
\newblock In \emph{Proceedings of the 61st Annual Meeting of the Association
  for Computational Linguistics (Volume 1: Long Papers)}, pages 36--50.

\bibitem[{Dale et~al.(2023{\natexlab{b}})Dale, Voita, Lam, Hansanti, Ropers,
  Kalbassi, Gao, Barrault, and Costa-juss{\`a}}]{dale2023halomi}
David Dale, Elena Voita, Janice Lam, Prangthip Hansanti, Christophe Ropers,
  Elahe Kalbassi, Cynthia Gao, Loic Barrault, and Marta Costa-juss{\`a}.
  2023{\natexlab{b}}.
\newblock \href {https://aclanthology.org/2023.emnlp-main.42} {{H}al{O}mi: A
  manually annotated benchmark for multilingual hallucination and omission
  detection in machine translation}.
\newblock In \emph{Proceedings of the 2023 Conference on Empirical Methods in
  Natural Language Processing}, pages 638--653.

\bibitem[{Devlin et~al.(2019)Devlin, Chang, Lee, and
  Toutanova}]{devlin2019bert}
Jacob Devlin, Ming-Wei Chang, Kenton Lee, and Kristina Toutanova. 2019.
\newblock \href {https://aclanthology.org/N19-1423} {{BERT}: Pre-training of
  deep bidirectional transformers for language understanding}.
\newblock In \emph{Proceedings of the 2019 Conference of the North American
  Chapter of the Association for Computational Linguistics: Human Language
  Technologies, Volume 1 (Long and Short Papers)}, pages 4171--4186.

\bibitem[{Dou and Neubig(2021)}]{dou2021word}
Zi-Yi Dou and Graham Neubig. 2021.
\newblock \href {https://aclanthology.org/2021.eacl-main.181} {Word alignment
  by fine-tuning embeddings on parallel corpora}.
\newblock In \emph{Proceedings of the 16th Conference of the European Chapter
  of the Association for Computational Linguistics: Main Volume}, pages
  2112--2128.

\bibitem[{Duquenne et~al.(2023)Duquenne, Schwenk, and
  Sagot}]{duquenne2308sonar}
Paul-Ambroise Duquenne, Holger Schwenk, and Benoit Sagot. 2023.
\newblock \href {https://arxiv.org/abs/2308.11466} {Sonar: {S}entence-level
  multimodal and language-agnostic representations}.
\newblock \emph{ArXiv Preprint}.

\bibitem[{Feng et~al.(2022)Feng, Yang, Cer, Arivazhagan, and
  Wang}]{feng2022language}
Fangxiaoyu Feng, Yinfei Yang, Daniel Cer, Naveen Arivazhagan, and Wei Wang.
  2022.
\newblock \href {https://aclanthology.org/2022.acl-long.62} {Language-agnostic
  bert sentence embedding}.
\newblock In \emph{Proceedings of the 60th Annual Meeting of the Association
  for Computational Linguistics (Volume 1: Long Papers)}, pages 878--891.

\bibitem[{Ferrando et~al.(2022)Ferrando, G{\'a}llego, Alastruey, Escolano, and
  Costa-juss{\`a}}]{ferrando2022towards}
Javier Ferrando, Gerard~I G{\'a}llego, Belen Alastruey, Carlos Escolano, and
  Marta~R Costa-juss{\`a}. 2022.
\newblock \href {https://aclanthology.org/2022.emnlp-main.599} {Towards opening
  the black box of neural machine translation: Source and target
  interpretations of the transformer}.
\newblock In \emph{Proceedings of the 2022 Conference on Empirical Methods in
  Natural Language Processing}, pages 8756--8769.

\bibitem[{Ghader and Monz(2017)}]{ghader2017does}
Hamidreza Ghader and Christof Monz. 2017.
\newblock \href {https://aclanthology.org/I17-1004} {What does attention in
  neural machine translation pay attention to?}
\newblock In \emph{Proceedings of the Eighth International Joint Conference on
  Natural Language Processing (Volume 1: Long Papers)}, pages 30--39.

\bibitem[{Guerreiro et~al.(2023{\natexlab{a}})Guerreiro, Alves, Waldendorf,
  Haddow, Birch, Colombo, and Martins}]{guerreiro2023hallucinations}
Nuno~M Guerreiro, Duarte~M Alves, Jonas Waldendorf, Barry Haddow, Alexandra
  Birch, Pierre Colombo, and Andr{\'e}~FT Martins. 2023{\natexlab{a}}.
\newblock \href
  {https://direct.mit.edu/tacl/article/doi/10.1162/tacl_a_00615/118716/Hallucinations-in-Large-Multilingual-Translation}
  {Hallucinations in large multilingual translation models}.
\newblock \emph{Transactions of the Association for Computational Linguistics},
  11:1500--1517.

\bibitem[{Guerreiro et~al.(2023{\natexlab{b}})Guerreiro, Colombo, Piantanida,
  and Martins}]{guerreiro2023optimal}
Nuno~M. Guerreiro, Pierre Colombo, Pablo Piantanida, and Andr{\'e} Martins.
  2023{\natexlab{b}}.
\newblock \href {https://aclanthology.org/2023.acl-long.770} {Optimal transport
  for unsupervised hallucination detection in neural machine translation}.
\newblock In \emph{Proceedings of the 61st Annual Meeting of the Association
  for Computational Linguistics (Volume 1: Long Papers)}, pages 13766--13784.

\bibitem[{Guerreiro et~al.(2023{\natexlab{c}})Guerreiro, Voita, and
  Martins}]{guerreiro2023looking}
Nuno~M Guerreiro, Elena Voita, and Andr{\'e}~FT Martins. 2023{\natexlab{c}}.
\newblock \href {https://aclanthology.org/2023.eacl-main.75/} {Looking for a
  needle in a haystack: A comprehensive study of hallucinations in neural
  machine translation}.
\newblock In \emph{Proceedings of the 17th Conference of the European Chapter
  of the Association for Computational Linguistics}, pages 1059--1075.

\bibitem[{Heffernan et~al.(2022)Heffernan, {\c{C}}elebi, and
  Schwenk}]{heffernan2022bitext}
Kevin Heffernan, Onur {\c{C}}elebi, and Holger Schwenk. 2022.
\newblock \href {https://aclanthology.org/2022.findings-emnlp.154} {Bitext
  mining using distilled sentence representations for low-resource languages}.
\newblock In \emph{Findings of the Association for Computational Linguistics:
  EMNLP 2022}, pages 2101--2112.

\bibitem[{Huang et~al.(2016)Huang, Guo, Kusner, Sun, Sha, and
  Weinberger}]{huang2016supervised}
Gao Huang, Chuan Guo, Matt~J Kusner, Yu~Sun, Fei Sha, and Kilian~Q Weinberger.
  2016.
\newblock \href
  {http://papers.neurips.cc/paper/6139-supervised-word-movers-distance.pdf}
  {Supervised word mover's distance}.
\newblock \emph{Advances in Neural Information Processing Systems}, 29.

\bibitem[{Kantorovich(1942)}]{Kantorovich1942}
L.~V. Kantorovich. 1942.
\newblock \href
  {https://www.math.toronto.edu/~mccann/assignments/477/Kantorovich42.pdf} {On
  the translocation of masses}.
\newblock \emph{Doklady Akademii Nauk}, 37(2):227--229.

\bibitem[{Koehn(2005)}]{koehn2005europarl}
Philipp Koehn. 2005.
\newblock \href {https://aclanthology.org/2005.mtsummit-papers.11/} {Europarl:
  A parallel corpus for statistical machine translation}.
\newblock In \emph{Proceedings of Machine Translation Summit X: Papers}, pages
  79--86.

\bibitem[{Koehn et~al.(2005)Koehn, Martin, Mihalcea, Monz, and
  Pedersen}]{koehn2005proceedings}
Philipp Koehn, Joel Martin, Rada Mihalcea, Christof Monz, and Ted Pedersen.
  2005.
\newblock \href {https://aclanthology.org/W05-0800} {Proceedings of the acl
  workshop on building and using parallel texts}.
\newblock In \emph{Proceedings of the ACL Workshop on Building and Using
  Parallel Texts}.

\bibitem[{Kuhn(1955)}]{kuhn1955hungarian}
Harold~W Kuhn. 1955.
\newblock \href {https://doi.org/10.1002/nav.3800020109} {The hungarian method
  for the assignment problem}.
\newblock \emph{Naval Research Logistics Quarterly}, 2(1-2):83--97.

\bibitem[{Lan et~al.(2021)Lan, Jiang, and Xu}]{lan2021neural}
Wuwei Lan, Chao Jiang, and Wei Xu. 2021.
\newblock \href {https://aclanthology.org/2021.acl-long.531} {Neural
  semi-markov crf for monolingual word alignment}.
\newblock In \emph{Proceedings of the 59th Annual Meeting of the Association
  for Computational Linguistics and the 11th International Joint Conference on
  Natural Language Processing (Volume 1: Long Papers)}, pages 6815--6828.

\bibitem[{Lee et~al.(2022)Lee, Lee, Jang, and Yu}]{lee2022toward}
Seonghyeon Lee, Dongha Lee, Seongbo Jang, and Hwanjo Yu. 2022.
\newblock \href {https://aclanthology.org/2022.acl-long.412} {Toward
  interpretable semantic textual similarity via optimal transport-based
  contrastive sentence learning}.
\newblock In \emph{Proceedings of the 60th Annual Meeting of the Association
  for Computational Linguistics (Volume 1: Long Papers)}, pages 5969--5979.

\bibitem[{Mare{\v{c}}ek(2011)}]{marevcek2011automatic}
David Mare{\v{c}}ek. 2011.
\newblock \href
  {https://ufal.mff.cuni.cz/pcedt3.0/pubs/Marecek2008_diplomka.pdf} {Automatic
  alignment of tectogrammatical trees from czech-english parallel corpus}.
\newblock \emph{Univerzita Karlova, Matematicko-fyzik{\'a}ln{\'\i} fakulta}.

\bibitem[{Mihalcea and Pedersen(2003)}]{mihalcea2003evaluation}
Rada Mihalcea and Ted Pedersen. 2003.
\newblock \href {https://aclanthology.org/W03-0301} {An evaluation exercise for
  word alignment}.
\newblock In \emph{Proceedings of the HLT-NAACL 2003 Workshop on Building and
  using parallel texts: data driven machine translation and beyond}, pages
  1--10.

\bibitem[{Monge(1781)}]{Monge1781}
Gaspard Monge. 1781.
\newblock \href {https://cir.nii.ac.jp/crid/1572261550791499008} {Mémoire sur
  la théorie des déblais et des remblais}.
\newblock \emph{Histoire de l’Académie Royale des Sciences de Paris}.

\bibitem[{M{\"u}ller and Sennrich(2021)}]{muller2021understanding}
Mathias M{\"u}ller and Rico Sennrich. 2021.
\newblock \href {https://aclanthology.org/2021.acl-long.21} {Understanding the
  properties of minimum bayes risk decoding in neural machine translation}.
\newblock In \emph{Proceedings of the 59th Annual Meeting of the Association
  for Computational Linguistics and the 11th International Joint Conference on
  Natural Language Processing (Volume 1: Long Papers)}, pages 259--272.

\bibitem[{Mysore et~al.(2022)Mysore, Cohan, and Hope}]{mysore2022multi}
Sheshera Mysore, Arman Cohan, and Tom Hope. 2022.
\newblock \href {https://aclanthology.org/2022.naacl-main.331} {Multi-vector
  models with textual guidance for fine-grained scientific document
  similarity}.
\newblock In \emph{Proceedings of the 2022 Conference of the North American
  Chapter of the Association for Computational Linguistics: Human Language
  Technologies}, pages 4453--4470.

\bibitem[{Nagata et~al.(2020)Nagata, Chousa, and
  Nishino}]{nagata2020supervised}
Masaaki Nagata, Katsuki Chousa, and Masaaki Nishino. 2020.
\newblock \href {https://aclanthology.org/2020.emnlp-main.41} {A supervised
  word alignment method based on cross-language span prediction using
  multilingual bert}.
\newblock In \emph{Proceedings of the 2020 Conference on Empirical Methods in
  Natural Language Processing (EMNLP)}, pages 555--565.

\bibitem[{Och and Ney(2000)}]{och2000improved}
Franz~Josef Och and Hermann Ney. 2000.
\newblock \href {https://aclanthology.org/P00-1056} {Improved statistical
  alignment models}.
\newblock In \emph{Proceedings of the 38th Annual Meeting of the Association
  for Computational Linguistics}, pages 440--447.

\bibitem[{Och and Ney(2003)}]{och2003systematic}
Franz~Josef Och and Hermann Ney. 2003.
\newblock \href
  {https://www.mitpressjournals.org/doi/abs/10.1162/089120103321337421} {A
  systematic comparison of various statistical alignment models}.
\newblock \emph{Computational linguistics}, 29(1):19--51.

\bibitem[{{\"O}stling and Tiedemann(2016)}]{ostling2016efficient}
Robert {\"O}stling and J{\"o}rg Tiedemann. 2016.
\newblock \href {https://ufal.mff.cuni.cz/pbml/106/art-ostling-tiedemann.pdf}
  {Efficient word alignment with markov chain monte carlo}.
\newblock \emph{The Prague Bulletin of Mathematical Linguistics}.

\bibitem[{Paszke et~al.(2019)Paszke, Gross, Massa, Lerer, Bradbury, Chanan,
  Killeen, Lin, Gimelshein, Antiga et~al.}]{paszke2019pytorch}
Adam Paszke, Sam Gross, Francisco Massa, Adam Lerer, James Bradbury, Gregory
  Chanan, Trevor Killeen, Zeming Lin, Natalia Gimelshein, Luca Antiga, et~al.
  2019.
\newblock \href {https://arxiv.org/abs/1912.01703} {Py{T}orch: {A}n imperative
  style, high-performance deep learning library}.
\newblock \emph{Advances in Neural Information Processing Systems}, 32.

\bibitem[{Peyr{\'e} et~al.(2016)Peyr{\'e}, Cuturi, and
  Solomon}]{peyre2016gromov}
Gabriel Peyr{\'e}, Marco Cuturi, and Justin Solomon. 2016.
\newblock \href {http://proceedings.mlr.press/v48/peyre16.html}
  {Gromov-wasserstein averaging of kernel and distance matrices}.
\newblock In \emph{International Conference on Machine Learning}, pages
  2664--2672.

\bibitem[{Raunak et~al.(2021)Raunak, Menezes, and
  Junczys-Dowmunt}]{raunak2021curious}
Vikas Raunak, Arul Menezes, and Marcin Junczys-Dowmunt. 2021.
\newblock \href {https://aclanthology.org/2021.naacl-main.92} {The curious case
  of hallucinations in neural machine translation}.
\newblock In \emph{Proceedings of the 2021 Conference of the North American
  Chapter of the Association for Computational Linguistics: Human Language
  Technologies}, pages 1172--1183.

\bibitem[{Rei et~al.(2020)Rei, Stewart, Farinha, and Lavie}]{rei2020comet}
Ricardo Rei, Craig Stewart, Ana~C Farinha, and Alon Lavie. 2020.
\newblock \href {https://aclanthology.org/2020.emnlp-main.213} {{COMET}: A
  neural framework for mt evaluation}.
\newblock In \emph{Proceedings of the 2020 Conference on Empirical Methods in
  Natural Language Processing (EMNLP)}, pages 2685--2702.

\bibitem[{Sabet et~al.(2020)Sabet, Dufter, Yvon, and
  Sch{\"u}tze}]{sabet2020simalign}
Masoud~Jalili Sabet, Philipp Dufter, Fran{\c{c}}ois Yvon, and Hinrich
  Sch{\"u}tze. 2020.
\newblock \href {https://aclanthology.org/2020.findings-emnlp.147} {Simalign:
  High quality word alignments without parallel training data using static and
  contextualized embeddings}.
\newblock In \emph{Findings of the Association for Computational Linguistics:
  EMNLP 2020}, pages 1627--1643.

\bibitem[{Tavakoli and Faili(2014)}]{tavakoli2014phrase}
Leila Tavakoli and Heshaam Faili. 2014.
\newblock \href {http://www.ijict.itrc.ac.ir/article-1-138-en.html} {Phrase
  alignments in parallel corpus using bootstrapping approach}.
\newblock \emph{International Journal of Information and Communication
  Technology Research}.

\bibitem[{Villani(2009)}]{villani2009optimal}
C{\'e}dric Villani. 2009.
\newblock \href {https://link.springer.com/book/10.1007/978-3-540-71050-9}
  {\emph{Optimal transport: {O}ld and new}}, volume 338.
\newblock Springer.

\bibitem[{Zenkel et~al.(2020)Zenkel, Wuebker, and DeNero}]{zenkel2020end}
Thomas Zenkel, Joern Wuebker, and John DeNero. 2020.
\newblock \href {https://aclanthology.org/2020.acl-main.146} {End-to-end neural
  word alignment outperforms giza++}.
\newblock In \emph{Proceedings of the 58th Annual Meeting of the Association
  for Computational Linguistics}, pages 1605--1617.

\bibitem[{Zhang and van Genabith(2021)}]{zhang2021bidirectional}
Jingyi Zhang and Josef van Genabith. 2021.
\newblock \href {https://aclanthology.org/2021.acl-long.24} {A bidirectional
  transformer based alignment model for unsupervised word alignment}.
\newblock In \emph{Proceedings of the 59th Annual Meeting of the Association
  for Computational Linguistics and the 11th International Joint Conference on
  Natural Language Processing (Volume 1: Long Papers)}, pages 283--292.

\bibitem[{Zhou et~al.(2021)Zhou, Neubig, Gu, Diab, Guzm{\'a}n, Zettlemoyer, and
  Ghazvininejad}]{zhou2021detecting}
Chunting Zhou, Graham Neubig, Jiatao Gu, Mona Diab, Francisco Guzm{\'a}n, Luke
  Zettlemoyer, and Marjan Ghazvininejad. 2021.
\newblock \href {https://aclanthology.org/2021.findings-acl.120} {Detecting
  hallucinated content in conditional neural sequence generation}.
\newblock In \emph{Findings of the Association for Computational Linguistics:
  ACL-IJCNLP 2021}, pages 1393--1404.

\end{thebibliography}

\appendix

\section{Equidistant Vector}
\label{app:math}
Assume that we have vectors $\bm  e_1, \dots, \bm  e_n$ in $\mathbb{R}^D$. Here we discuss how to find an equidistant vector $\bm e^{(\varnothing)}$ with respect to cosine distance. For that we need to solve a system of $N-1$ equations:
$$\textbf{dist}(e^{\varnothing}, \bm  e_1) = \textbf{dist}(e^{\varnothing}, \bm  e_j)  $$
for all $j=2, \dots, N$. 

Recall the definition of the cosine distance:
$$\textbf{dist}(\bm x, \bm y) = 1 - \frac{\bm  x \cdot \bm  y}{\|\bm  x\| \|\bm  y\|} .$$

Hence, the system of equations we need to solve  is equivalent to the system of linear equations:
$$\frac{\bm  e^{\varnothing} \cdot \bm  e_1}{\|\bm  e^{\varnothing}\| \|\bm  e_1\|} = \frac{\bm  e^{\varnothing} \cdot \bm  e_j}{\|\bm  e^{\varnothing}\| \|\bm  e_j\|}. $$

Slightly rewriting it, we get:
$$\bm  e^{\varnothing} \cdot \left( \frac{ \bm  e_1}{ \|\bm  e_1\|} - \frac{ \bm  e_j}{ \|\bm  e_j\|} \right) = 0 .$$

Note that usually for our word alignment problems $N < D$, hence this system always has a solution. 
However, we want to find the minimum possible equal distance so that the NULL vector is meaningful for the OT task. For that, assume that  $\bm e^{(\varnothing)}$ lies in the span of the given vectors: $\bm e^{\varnothing} = \sum_{k=1}^N a_k \bm e_k$. Inserting it in the previous system, we obtain a new homogeneous system on the coefficients $a_k$:
$$ \bm E \bm a = \bm 0 , $$
where $E_{jk} = \bm  e_k \cdot \left( \frac{ \bm  e_1}{ \|\bm  e_1\|} - \frac{ \bm  e_j}{ \|\bm  e_j\|} \right)$ is a $(N-1)\times N$ matrix. For general positions of word representation vectors, the kernel of $\bm E$ is one-dimensional. Projection onto it could be computed as $\bm I_N - \bm E^{+}\bm E$, where $\bm E^{+}$ is Moore-Penrose inverse and $\bm I_N$ is $N \times N$ identity matrix \cite{Campbell1979GeneralizedIO}. In this case $\bm a$ will be the basis vector for the kernel subspace. After finding $\bm a$, one can construct $e^{\varnothing} =  \sum_{k=1}^N a_k \bm e_k$ and compute $d_{\text{min}} = \textbf{dist}(e^{\varnothing}, e_1)$. Note that there is no equidistant vector with a smaller distance than $d_{\text{min}}$.

\section{Additional Results}
\label{app:Results}
We present the comparisons of different word embeddings and alignment methods in Table~\ref{tab:abl_table}.

\begin{table*}[!ht] 

\centering
 \renewcommand{\arraystretch}{1.25}
\resizebox{1\textwidth}{!}{%
\begin{tabular}{l@{\hspace{0.75\tabcolsep}}l@{\hspace{0.75\tabcolsep}}|c@{\hspace{0.75\tabcolsep}}c@{\hspace{0.75\tabcolsep}}c@{\hspace{0.75\tabcolsep}}c@{\hspace{0.75\tabcolsep}}c@{\hspace{0.75\tabcolsep}}c@{\hspace{0.75\tabcolsep}}c@{\hspace{0.75\tabcolsep}}|c@{\hspace{0.75\tabcolsep}}c@{\hspace{0.75\tabcolsep}}c@{\hspace{0.75\tabcolsep}}c@{\hspace{0.75\tabcolsep}}c@{\hspace{0.75\tabcolsep}}c@{\hspace{0.75\tabcolsep}}c}  
\toprule  
\multirow{2}{*}{\textbf{\colorbox{blue!0}{Source}}} &
\multirow{2}{*}{\textbf{\colorbox{blue!0}{Lang}}} & 
\multicolumn{7}{c|}{\textit{Hallucination}} & \multicolumn{7}{c}{\textit{Omission}} \\ 
& & 

\textbf{\colorbox{purple!20}{{$\mathcal{W}1$}}}  &
\textbf{\colorbox{purple!20}{{$\mathcal{W}2$}}}&
\textbf{\colorbox{purple!20}{{$\mathcal{W}3$}}} &
\textbf{\colorbox{purple!20}{{$\mathcal{W}4$}}}&
\textbf{\colorbox{yellow!20}{{Our}}} &
\textbf{\colorbox{green!30}{{$\mathcal{R}1$}}} &
\textbf{\colorbox{green!30}{{$\mathcal{W}2$}}} &

\textbf{\colorbox{purple!20}{{$\mathcal{W}1$}}}  &
\textbf{\colorbox{purple!20}{{$\mathcal{W}2$}}}&
\textbf{\colorbox{purple!20}{{$\mathcal{W}3$}}} &
\textbf{\colorbox{purple!20}{{$\mathcal{W}4$}}}&
\textbf{\colorbox{yellow!20}{{Our}}} &
\textbf{\colorbox{green!30}{{$\mathcal{R}1$}}} &
\textbf{\colorbox{green!30}{{$\mathcal{W}2$}}}
\\
\midrule 

\multicolumn{1}{l}{\multirow{10}{*}{\begin{sideways}\specialcell{High Resource}\end{sideways}}}      & en-ar &       33 &       64 &         66 &  65 & \textbf{93} &         85 &         88 &       56 &         80 &         82 &         78 &         79 & \textbf{84} &         81 \\
 & ar-en &       49 &       83 &         82 &  69 & \textbf{90} &         85 &         88 &       71 & \textbf{74} &         67 &         65 &         70 &         68 &         63 \\
 & en-ru &       31 &       74 &         75 &  71 &         91 &         87 & \textbf{94} &       74 &         83 &         82 &         76 & \textbf{85} &         83 &         76 \\
 & ru-en &       51 &       81 &         80 &  50 & \textbf{99} &         97 &         91 &       75 &         87 & \textbf{90} &         80 &         89 &         85 &         81 \\
 & en-es &       54 &       88 &         86 &  77 & \textbf{88} &         85 &         84 &       64 &         88 & \textbf{89} &         70 &         87 &         88 &         89 \\
 & es-en &       60 &       83 & \textbf{88} &  67 &         87 &         83 &         83 &       69 & \textbf{79} &         75 &         65 &         73 &         73 &         68 \\
 & en-de &       40 &       86 &         86 &  52 & \textbf{88} &         81 &         83 &       65 &         78 &         81 &         78 &         81 &         82 & \textbf{84} \\
 & de-en &       62 &       94 &         95 &  67 & \textbf{96} &         92 &         94 &       75 &         79 & \textbf{80} &         64 &         74 &         75 &         75 \\
 & en-zh &       39 &       72 &         72 &  59 & \textbf{88} &         87 &         85 &       84 &         92 & \textbf{94} &         79 &         92 &         93 &         88 \\
 & zh-en &       63 &       85 &         83 &  53 &         86 & \textbf{87} &         77 &       61 & \textbf{80} &         77 &         72 &         77 &         79 &         66 \\

\midrule
\multicolumn{2}{l|}{\textbf{Avg. High Resouce}} &       47 &       81 &         81 &  64 & \textbf{91} &         87 &         88 &       70 & \textbf{82} &         82 &         73 &         81 &         81 &         78 \\

\midrule

\multicolumn{1}{l}{\multirow{6}{*}{\begin{sideways}\specialcell{Low Resource}\end{sideways}}}    & en-ks &       51 &       62 &         65 &  47 & \textbf{78} &         62 &         52 &       62 &         84 &         78 &         46 & \textbf{84} &         83 &         71 \\
 & ks-en &       51 &       53 &         51 &  44 & \textbf{56} &         56 &         56 &       56 &         66 & \textbf{71} &         49 &         70 &         61 &         68 \\
& en-mni &       55 &       56 &         56 &  51 & \textbf{59} &         50 &         52 &       55 &         63 &         67 &         65 & \textbf{69} &         60 &         49 \\
& mni-en &       57 &       59 &         56 &  56 & \textbf{64} &         50 &         52 &       66 &         67 &         74 &         61 & \textbf{77} &         65 &         63 \\
 & en-yo &       55 &       77 &         74 &  60 & \textbf{80} &         67 &         57 &       69 &         82 &         85 &         75 & \textbf{86} &         76 &         72 \\
 & yo-en &       57 &       59 &         59 &  52 & \textbf{65} &         60 &         45 &       61 &         66 &         68 &         62 &         68 & \textbf{69} &         59 \\

\midrule
\multicolumn{2}{l|}{\textbf{Avg. Low Resouce}} &       55 &       61 &         60 &  52 & \textbf{67} &         56 &         53 &       62 &         71 &         74 &         60 & \textbf{76} &         69 &         64 \\

\midrule
\multirow{2}{*}{\parbox{1mm}{Zero-Shot}}    & yo-es &       51 &       54 &         54 &  47 & \textbf{54} &         48 &         45 &       72 &         76 &         77 & \textbf{80} &         76 &         76 &         73 \\
 & es-yo &       62 &       62 &         63 &  56 & \textbf{65} &         57 &         56 &       59 &         80 & \textbf{85} &         53 &         80 &         76 &         72 \\

\midrule
\multicolumn{2}{l|}{\textbf{Avg. Zero-Shot}} &       56 &       57 &         57 &  52 & \textbf{59} &         53 &         51 &       66 &         78 & \textbf{81} &         67 &         78 &         76 &         72 \\

\midrule
\multicolumn{2}{l|}{\textbf{Avg. HalOmi}} &       51 &       72 &         72 &  57 & \textbf{79} &         73 &         71 &       66 &         78 &         \textbf{79} &         68 & \textbf{79} &         76 &         72 \\
\bottomrule  

\end{tabular} 
}

\caption{Methods performances (ROC AUC) on hallucination and omission detection across HalOmi's high-resource, low-resource, and zero-shot language-pair clusters. Bold entries describe the best results among all methods. \method{} refer to using \method{} word aligner with LABSE embeddings. \colorbox{purple!20}{{$\mathcal{W}1$, $\mathcal{W}2$, $\mathcal{W}3$, $\mathcal{W}4$}}: SimAlign-Itermax~\cite{sabet2020simalign}, PMIAlign~\cite{azadi2023pmi}, OT~\cite{azadi2023pmi}, POT~\cite{arase2023unbalanced} word aligners with LaBSE embeddings. \colorbox{green!20}{{$\mathcal{R}1$, $\mathcal{R}2$}}: \method{} with embeddings obtained from m\bert{}~\cite{devlin2019bert} and XLMR~\cite{conneau2020unsupervised}.}   
\label{tab:abl_table}

\end{table*} 

\end{document}